\pdfoutput=1
\documentclass[11pt]{article}

\usepackage{EMNLP2022}

\usepackage{times}
\usepackage{latexsym}

\usepackage[T1]{fontenc}
\usepackage[utf8]{inputenc}

\usepackage{microtype}

\usepackage{inconsolata}

\usepackage{amsmath}
\usepackage{booktabs}
\usepackage{xspace}
\usepackage{pifont}
\usepackage{threeparttable}
\usepackage{tikz}
\usepackage{graphicx} 
\usepackage{pgfplots}
\usepackage{multirow}
\RequirePackage{algorithm}
\RequirePackage{algorithmic}
\usepackage{enumitem}

\newcommand{\method}{\textsc{ReaRev}\xspace}

\usepackage{amsmath,amsfonts,bm}

\def\Figref#1{Figure~\ref{#1}}

\def\Secref#1{Section~\ref{#1}}
\def\eqref#1{~(\ref{#1})}
\def\Eqref#1{Eq.(\ref{#1})}
\def\1{\bm{1}}

\def\vb{{\bm{b}}}
\def\vc{{\bm{c}}}

\def\vg{{\bm{g}}}
\def\vh{{\bm{h}}}
\def\vi{{\bm{i}}}

\def\vm{{\bm{m}}}

\def\vp{{\bm{p}}}
\def\vq{{\bm{q}}}
\def\vr{{\bm{r}}}

\def\vw{{\bm{w}}}

\def\mH{{\bm{H}}}

\def\mW{{\bm{W}}}

\DeclareMathAlphabet{\mathsfit}{\encodingdefault}{\sfdefault}{m}{sl}
\SetMathAlphabet{\mathsfit}{bold}{\encodingdefault}{\sfdefault}{bx}{n}

\def\gF{{\mathcal{F}}}
\def\gG{{\mathcal{G}}}

\def\gN{{\mathcal{N}}}
\def\gO{{\mathcal{O}}}

\def\gR{{\mathcal{R}}}

\def\gV{{\mathcal{V}}}

\def\sR{{\mathbb{R}}}

\newcommand{\softmax}{\mathrm{softmax}}

\title{ReaRev: Adaptive Reasoning for Question Answering over Knowledge
Graphs }

\author{Costas Mavromatis \and George Karypis\\
        University of Minnesota, USA \\
        \texttt{ \{mavro016, karypis\}@umn.edu} \\}

\begin{document}
\maketitle
\begin{abstract}
Knowledge Graph Question Answering (KGQA) involves retrieving entities as answers from a Knowledge Graph (KG) using natural language queries. The challenge is to learn to reason over question-relevant KG facts that traverse KG entities and  lead to the question answers. 
%In the weakly supervised setting, only question-answer pairs are given as supervision.
%Existing state-of-the-art methods rely on graph-based neural networks (GNNs) to selectively aggregate information from question-relevant facts of the KG. 
To facilitate reasoning, the question is decoded into instructions, which are dense question representations used to guide the KG traversals. However, if the derived instructions do not exactly match the underlying KG information,  they may lead to reasoning under irrelevant context.
Our method, termed \method, introduces a new way to KGQA reasoning with respect
to both instruction decoding and execution. To improve instruction decoding, we perform reasoning in an adaptive manner, where KG-aware information is used to iteratively update the initial instructions. To improve instruction execution, we emulate breadth-first search (BFS) with graph neural networks (GNNs). The BFS strategy treats the instructions as a set and allows our method to decide on their execution order on the fly. Experimental results on three KGQA benchmarks demonstrate the  \method's effectiveness compared with previous state-of-the-art, especially when the KG is incomplete or when we tackle complex questions. Our code is publicly available at \url{https://github.com/cmavro/ReaRev_KGQA}.
% Using question-answer pairs, existing methods learn natural question representations (instructions) to guide the reasoning over KGs, which is typically performed with graph neural networks (GNNs). The challenge is that ground-truth reasoning paths to answers are seldom available, which does not allow evaluating intermediate reasoning steps. In this work, we propose to tackle this challenge by reasoning in an adaptive multistep fashion to improve the generated reasoning paths. Our method \method employs a specialized GNN module  (\methodg) that emulates either the breadth-first-search (BFS) or the depth-first-search (DFS) to support parallel or deep sequential reasoning, respectively.
% In both cases, \method repeats the reasoning process for multiple iterations, so that it adaptively explores alternative reasoning paths in an end-to-end manner. Experimental results on three KGQA benchmarks demonstrate the effectiveness of \method compared with previous state-of-the-art. 
\end{abstract}

\section{Introduction}

A knowledge graph (KG) is a relational graph that contains a set of known facts. These facts are represented as tuples \textit{(subject, relation, object)}, where the subject and object are KG entities that link with the given relation. The knowledge graph question-answering (KGQA) task takes as input a natural language question and returns a set of KG entities as the answer. 
%KGQA solutions involve transforming the question to useful instructions, which can be executed over the KG to arrive at the answers, inducing a reasoning path. %This leads to executing a specific reasoning path over the KG such as following specific relations. 
% A line of work~\cite{berant2013semantic,yih2015semantic,luo2018knowledge,lan2020query} parses the questions into executable queries, e.g., SPARQL queries, whose execution leads to KG traversals that return the answers. However, these methods require ground-truth executable queries or predefined question templates during training, which are costly and hard to obtain.
In the weakly-supervised KGQA setting~\cite{berant2013semantic}, the only available supervisions during learning are question-answer pairs, e.g.,  ``Q: \textit{Who is the director of Pulp Fiction}? A: \textit{Q. Tarantino}''. Due to labelling costs, the ground-truth KG traversals, such as the path $ \textit{Pulp Fiction} \xrightarrow{\text{director}} \textit{Q. Tarantino}$, whose execution leads to the answers, are seldom available.
\begin{figure}[t]
\centering
\resizebox{0.85\columnwidth}{!}{\tikzset{every picture/.style={line width=0.75pt}} %set default line width to 0.75pt        

\begin{tikzpicture}[x=0.75pt,y=0.60pt,yscale=-1,xscale=1]
%uncomment if require: \path (0,311); %set diagram left start at 0, and has height of 311

%Shape: Circle [id:dp4202883122823482] 
\draw  [fill={rgb, 255:red, 245; green, 166; blue, 35 }  ,fill opacity=1 ] (129,155) .. controls (129,150.58) and (132.58,147) .. (137,147) .. controls (141.42,147) and (145,150.58) .. (145,155) .. controls (145,159.42) and (141.42,163) .. (137,163) .. controls (132.58,163) and (129,159.42) .. (129,155) -- cycle ;
%Shape: Circle [id:dp9151299223492708] 
\draw  [fill={rgb, 255:red, 184; green, 233; blue, 134 }  ,fill opacity=1 ] (68,77) .. controls (68,72.58) and (71.58,69) .. (76,69) .. controls (80.42,69) and (84,72.58) .. (84,77) .. controls (84,81.42) and (80.42,85) .. (76,85) .. controls (71.58,85) and (68,81.42) .. (68,77) -- cycle ;
%Shape: Circle [id:dp5656175692891405] 
\draw  [fill={rgb, 255:red, 248; green, 231; blue, 28 }  ,fill opacity=1 ] (41,134) .. controls (41,129.58) and (44.58,126) .. (49,126) .. controls (53.42,126) and (57,129.58) .. (57,134) .. controls (57,138.42) and (53.42,142) .. (49,142) .. controls (44.58,142) and (41,138.42) .. (41,134) -- cycle ;
%Straight Lines [id:da25547590662477626] 
\draw    (165,115) -- (84,84) ;
%Straight Lines [id:da39033087448910875] 
\draw    (256,221) -- (222,221) ;
%Shape: Circle [id:dp5631483656704255] 
\draw  [fill={rgb, 255:red, 194; green, 19; blue, 254 }  ,fill opacity=1 ] (154,209) .. controls (154,204.58) and (157.58,201) .. (162,201) .. controls (166.42,201) and (170,204.58) .. (170,209) .. controls (170,213.42) and (166.42,217) .. (162,217) .. controls (157.58,217) and (154,213.42) .. (154,209) -- cycle ;
%Straight Lines [id:da7870880701785672] 
\draw    (84,84) -- (129,148) ;
%Shape: Circle [id:dp683554764680097] 
\draw  [fill={rgb, 255:red, 80; green, 227; blue, 194 }  ,fill opacity=1 ] (167,82) .. controls (167,77.58) and (170.58,74) .. (175,74) .. controls (179.42,74) and (183,77.58) .. (183,82) .. controls (183,86.42) and (179.42,90) .. (175,90) .. controls (170.58,90) and (167,86.42) .. (167,82) -- cycle ;
%Straight Lines [id:da33671786714874186] 
\draw    (64,187) -- (49,142) ;
%Shape: Circle [id:dp9954025773852062] 
\draw  [fill={rgb, 255:red, 144; green, 19; blue, 254 }  ,fill opacity=1 ] (56,195) .. controls (56,190.58) and (59.58,187) .. (64,187) .. controls (68.42,187) and (72,190.58) .. (72,195) .. controls (72,199.42) and (68.42,203) .. (64,203) .. controls (59.58,203) and (56,199.42) .. (56,195) -- cycle ;
%Straight Lines [id:da4831120759717997] 
\draw    (84,77) -- (167,82) ;
%Straight Lines [id:da9585425183700516] 
\draw    (162,201) -- (142,160) ;
%Shape: Circle [id:dp5013756819905524] 
\draw  [fill={rgb, 255:red, 208; green, 2; blue, 27 }  ,fill opacity=1 ] (164,123) .. controls (164,118.58) and (167.58,115) .. (172,115) .. controls (176.42,115) and (180,118.58) .. (180,123) .. controls (180,127.42) and (176.42,131) .. (172,131) .. controls (167.58,131) and (164,127.42) .. (164,123) -- cycle ;
%Straight Lines [id:da06400423001386835] 
\draw    (142,150) -- (164,123) ;
%Straight Lines [id:da006500669556685823] 
\draw    (49,126) -- (76,85) ;
%Curve Lines [id:da8766655855101597] 
\draw [color={rgb, 255:red, 126; green, 211; blue, 33 }  ,draw opacity=1 ][line width=1.5]  [dash pattern={on 5.63pt off 4.5pt}]  (81,93) .. controls (62.38,113.58) and (61.05,143.76) .. (63.82,184.49) ;
\draw [shift={(64,187)}, rotate = 265.91] [color={rgb, 255:red, 126; green, 211; blue, 33 }  ,draw opacity=1 ][line width=1.5]    (14.21,-4.28) .. controls (9.04,-1.82) and (4.3,-0.39) .. (0,0) .. controls (4.3,0.39) and (9.04,1.82) .. (14.21,4.28)   ;
\draw [shift={(81,93)}, rotate = 312.14] [color={rgb, 255:red, 126; green, 211; blue, 33 }  ,draw opacity=1 ][line width=1.5]    (0,6.71) -- (0,-6.71)   ;
%Curve Lines [id:da40901929579954577] 
\draw [color={rgb, 255:red, 126; green, 211; blue, 33 }  ,draw opacity=1 ][line width=1.5]  [dash pattern={on 5.63pt off 4.5pt}]  (84,84) .. controls (123,176.63) and (140.14,171.34) .. (153.02,206.23) ;
\draw [shift={(154,209)}, rotate = 251.11] [color={rgb, 255:red, 126; green, 211; blue, 33 }  ,draw opacity=1 ][line width=1.5]    (14.21,-4.28) .. controls (9.04,-1.82) and (4.3,-0.39) .. (0,0) .. controls (4.3,0.39) and (9.04,1.82) .. (14.21,4.28)   ;
\draw [shift={(84,84)}, rotate = 247.17] [color={rgb, 255:red, 126; green, 211; blue, 33 }  ,draw opacity=1 ][line width=1.5]    (0,6.71) -- (0,-6.71)   ;
%Straight Lines [id:da095527716295007] 
\draw [color={rgb, 255:red, 126; green, 211; blue, 33 }  ,draw opacity=1 ][line width=1.5]  [dash pattern={on 5.63pt off 4.5pt}]  (9,251) -- (44,251.92) ;
\draw [shift={(47,252)}, rotate = 181.51] [color={rgb, 255:red, 126; green, 211; blue, 33 }  ,draw opacity=1 ][line width=1.5]    (14.21,-4.28) .. controls (9.04,-1.82) and (4.3,-0.39) .. (0,0) .. controls (4.3,0.39) and (9.04,1.82) .. (14.21,4.28)   ;
%Straight Lines [id:da7179939481715667] 
\draw    (282,25) -- (282,42) ;
\draw [shift={(282,44)}, rotate = 270] [color={rgb, 255:red, 0; green, 0; blue, 0 }  ][line width=0.75]    (10.93,-3.29) .. controls (6.95,-1.4) and (3.31,-0.3) .. (0,0) .. controls (3.31,0.3) and (6.95,1.4) .. (10.93,3.29)   ;
%Shape: Rectangle [id:dp9785168567808706] 
\draw   (256,45) -- (331,45) -- (331,88) -- (256,88) -- cycle ;

% Text Node
\draw (149,150) node [anchor=north west][inner sep=0.75pt]  [font=\normalsize] [align=left] {\textit{{\normalsize Pulp Fiction}}};
% Text Node
\draw (2,48) node [anchor=north west][inner sep=0.75pt]  [font=\normalsize] [align=left] {\textit{{\normalsize Q. Tarantino}}};
% Text Node
\draw (2,140) node [anchor=north west][inner sep=0.75pt]  [font=\normalsize] [align=left] {\textit{{\normalsize Kill Bill}}};
% Text Node
\draw (2,3) node [anchor=north west][inner sep=0.75pt]  [font=\normalsize] [align=left] {\textbf{Q}: ``\textit{Which year are Q.Tarantino's movies directed?}''\\\textbf{A}: \ 2003, 1994};
% Text Node
\draw (260,213) node [anchor=north west][inner sep=0.75pt]  [font=\normalsize] [align=left] {KG links};
% Text Node
\draw (45,213) node [anchor=north west][inner sep=0.75pt]  [font=\normalsize] [align=left] {\textit{{\normalsize 2003}}};
% Text Node
\draw (20.5,94.5) node [anchor=north west][inner sep=0.75pt]   [align=left] {\textit{{\fontfamily{ptm}\normalsize direct}}};
% Text Node
\draw (172,59) node [anchor=north west][inner sep=0.75pt]  [font=\normalsize] [align=left] {\textit{{\normalsize 1963}}};
% Text Node
\draw (29,174) node [anchor=north west][inner sep=0.75pt]   [align=left] {\textit{{\fontfamily{ptm}\normalsize date}}};
% Text Node
\draw (116,60) node [anchor=north west][inner sep=0.75pt]   [align=left] {\textit{{\fontfamily{ptm}\normalsize date}}};
% Text Node
\draw (164,220) node [anchor=north west][inner sep=0.75pt]  [font=\normalsize] [align=left] {\textit{{\normalsize 1994}}};
% Text Node
\draw (127.5,82.5) node [anchor=north west][inner sep=0.75pt]   [align=left] {\textit{{\fontfamily{ptm}\normalsize cast}}};
% Text Node
\draw (157,173.5) node [anchor=north west][inner sep=0.75pt]   [align=left] {\textit{{\fontfamily{ptm}\normalsize aired on}}};
% Text Node
\draw (132,120) node [anchor=north west][inner sep=0.75pt]   [align=left] {\textit{{\fontfamily{ptm}\normalsize act}}};
% Text Node
\draw (185,98) node [anchor=north west][inner sep=0.75pt]  [font=\normalsize] [align=left] {\textit{{\normalsize U.Thurman}}};
% Text Node
\draw (55,244) node [anchor=north west][inner sep=0.75pt]  [font=\normalsize] [align=left] {correct KG traversals};
% Text Node
\draw (258,48) node [anchor=north west][inner sep=0.75pt]   [align=left] {\textbf{{\small instruction}}\\\textbf{{\small decoding}}};
% Text Node
\draw (258,91.4) node [anchor=north west][inner sep=0.75pt]    {$\left\{i^{( 1)} ,\ i^{( 2)}\right\}$};

% Text Node
\draw (78.5,124.5) node [anchor=north west][inner sep=0.75pt]   [align=left] {\textit{{\fontfamily{ptm}\normalsize direct}}};
% Text Node

% Text Node
\draw (10,265) node [anchor=north west][inner sep=0.75pt]  [font=\normalsize] [align=left] {Execution $i^{(1)} \xrightarrow{} i^{(2)}$  \ding{55} : Tarantino$\xrightarrow{\text{date}}$ \ding{55}};

\draw (10,290) node [anchor=north west][inner sep=0.75pt]  [font=\normalsize] [align=left] { Execution $i^{(2)} \xrightarrow{} i^{(1)}$  \ding{51} : Tarantino$\xrightarrow{ \text{direct}}$Kill Bill$\xrightarrow{\text{date}}$2003 \ding{51}};

\draw (10,315) node [anchor=north west][inner sep=0.75pt]  [font=\normalsize] [align=left] {Decoding \ding{55}: Tarantino$\xrightarrow{\text{direct}}\xrightarrow{\text{aired on}}$ \ding{55}};

\end{tikzpicture}}
\caption{The given question is decomposed into instructions $\{ i^{(1)}, i^{(2)}\}$ that are matched with relations \textit{date} and \textit{direct}, respectively. If the instructions are executed in an incorrect order ($i^{(1)} \xrightarrow{} i^{(2)}$), we cannot arrive at the answer. Moreover, if the instructions cannot be matched with the relation \textit{aired on} (right KG traversal), we cannot arrive at the second answer.}
\label{fig:ex}
\end{figure}
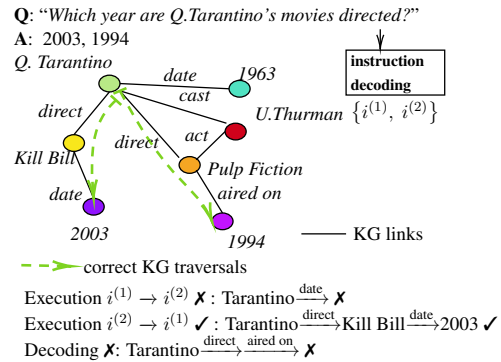

The KGQA problem involves two modules: (i) retrieving question-relevant KG facts, and (ii) reasoning over them to arrive at the answers. For the reasoning module, a general approach~\cite{lan2021survey} is to decode the question into dense representations (instructions) that guide the reasoning over the KG. The instructions are matched with one-hop KG facts (or relations) in an iterative manner, which induces KG traversals that lead to the answers. Instructions are usually generated by attending to different question's parts, e.g., ``\textit{...is the director...}''~\cite{qiu2020stepwise}.
KG traversals are  typically performed by utilizing powerful graph reasoners, such as graph neural networks (GNNs)~\cite{kipf2016semi, sun2018open}.

% Current state-of-the-art methods~\cite{sun2019pullnet,he2021improving,atzeni2021sqaler} tackle KGQA by improving the retrieval module to facilitate  semantic matching between instructions and KG facts. However, achieving accurate retrieval becomes challenging when the questions involve multiple facts (complex questions~\cite{talmor2018web}). Such questions involve complex KG traversals that cannot be trivially learned for retrieving \emph{all} necessary facts.
% This either limits these methods' performance gains~\cite{sun2019pullnet,he2021improving} or hinders their applicability~\cite{atzeni2021sqaler}.

The main challenge is that  question answering is performed over rich graph structures with possibly complex semantics, and thus, instruction decoding and execution is a key factor for KGQA. \Figref{fig:ex} shows an example where suboptimal initial instructions lead to incorrect KG traversals. While some methods~\cite{miller2016key,zhou2018interpretable,sun2018open,xu2019enhancing} attempt to improve the quality of the instructions, they  are mainly  designed  to tackle specific question types, such as 2-hop or 3-hop questions, or show poor performance for complex questions~\cite{sun2019pullnet}.

Our method termed \method (Reason \& Revise) introduces a new way to KGQA reasoning with respect to both instruction execution and decoding. To improve instruction execution, we do not use instructions in a pre-defined (possibly incorrect) order, but allow our method to decide on the execution order on the fly. We achieve this by emulating breadth-first search (BFS) with GNN reasoners. The BFS strategy treats the instructions as a set and the GNN decides which instructions to accept. To improve
instruction decoding, we reason over the KG to obtain KG-aware information and use this information to adapt the initial instructions. Then, we restart the reasoning with the new instructions that are conditioned  to the underlying KG semantics.
To the best of our knowledge, adaptive reasoning with emulating graph search algorithms with GNNs has not been previously proposed for KGQA. 

% is to emulate classical graph search algorithms, such as depth-first search, and explore as useful KG structure for each question as possible.  Our approach is to reason with each instruction for multiple hops (we are not limited to single hops), so that we induce multiple KG traversals. Then, we use the reasoning output of each instruction as KG-aware information to update the initial instruction decoding. This ensures that we reason under KG-relevant context. We repeat the above steps to iteratively explore additional KG traversals. We provide a motivating example in \Figref{fig:ex} behind our intuition.

% Second, we iteratively adapt the reasoning to the underlying structure to best reflect the KG semantics. 

% Our novel idea is to emulate classical graph search algorithms, such as depth-first search, to explore different KG traversals. We do this by reasoning with each instruction for multiple hops (search) to induce instruction-specific KG traversals. Moreover, we use the output of each instruction-specific reasoning as ....

We empirically show that \method performs effective reasoning over KGs and outperforms other state-of-the-art. For KGQA with complex questions, \method achieves improvement for 4.1 percentage points at Hits@1 over the best competing approach. 

Our contributions are summarized below:
\begin{itemize}[noitemsep]
    \item We improve instruction decoding via adaptive reasoning, which updates the instructions with KG-aware information. 
    \item We improve instruction execution  by emulating the breadth-first search algorithm with graph neural networks, which provides robustness to the instruction ordering.
    \item We achieve state-of-the-art (or nearly) performance on three widely used KGQA datasets: WebQuestions~\cite{yih2015semantic}, Complex WebQuestions~\cite{talmor2018web}, and MetaQA~\cite{zhang2018variational}.
\end{itemize}

\section{Related Work}

There are two mainstream approaches to solve KGQA: (i) parsing the question to executable KG queries like SPARQL,  and (ii) grounding question and KG representations to a common space for reasoning.

Regarding the first case, early methods~\cite{berant2013semantic,reddy2014large,bast2015more} rely on pre-defined question templates to synthesize queries, which  requires strong domain knowledge. Recent methods~\cite{yih2015semantic,abujabal2017automated,luo2018knowledge,bhutani2019learning,lan2019topic,lan2020query,qiu2020hierarchical,sun2021skeleton,das2021case} use deep learning techniques to automatically generate such executable queries. However, they need ground-truth executable queries as supervisions (which are costly to obtain) and their performance is limited when the KG has missing links (non-executable queries). 

Methods in the second category alleviate the need for ground-truth queries by learning natural language and KG representations to reason in a common space. These methods match question representations to KG facts~\cite{miller2016key,xu2019enhancing,atzeni2021sqaler} or to KG structure representations~\cite{zhang2018variational,han2021two,qiu2020stepwise}. More related to our work, GraftNet~\cite{sun2018open}, PullNet~\cite{sun2019pullnet}, and NSM~\cite{he2021improving} enhance the reasoning with graph-based reasoners. Our approach aims at improving the graph-based reasoning via adaptive instruction decoding and execution.

%Significant progress for KGQA without ground-truth queries has been made with deep learning based on graph neural networks (GNNs)
% \textbf{Beyond KGQA Static Reasoning}.
Researchers have also considered the problem of performing KGQA over incomplete graphs~\cite{min2013distant}, where important information is missing. These methods either rely on KG embeddings~\cite{saxena2020improving,ren2021lego} or on side information, such as text corpus~\cite{sun2018open,sun2019pullnet,xiong2019improving,han2020open}, to infer missing information. However, they offer marginal improvement over other methods when the KG is mostly complete. %Better improvement is observed by methods such as~\cite{das2017go,lin2018multi}, but only work with provided ground-truth executable queries. 

KGQA has also been adapted to specific domains, such as QA over temporal~\cite{mavromatis2021tempoqr,saxena2021question} and commonsense~\cite{speer2017conceptnet,talmor2019commonsenseqa,lin2019kagnet,feng2020scalable,yasunaga2021qa} KGs.

\section{Background}

%\subsection{Problem Setting}
% We provide the necessary background and notations. In this paper, we use the terms \emph{nodes} and \emph{entities} interchangeably. We denote vectors by bold lower-case letters and
% matrices by bold upper-case letters.

\subsection{KG}
A KG $\gG := (\gV, \gR, \gF)$ contains a set of entities $\gV$, a set of relations $\gR$, and a set of facts $\gF$.
Each fact $(v, r, v') \in \gF$ is a tuple where $v, v' \in \gV$ denote the subject and object entities, respectively, and $r \in \gR$ denotes the relation that holds between them. A KG is represented as a directed graph with $|\gR|$ relation types, where nodes $v$ and $v'$ connect with a directed relation  $r$ if $(v,r,v') \in \gF$. Nodes and relations are usually initialized with $d$-dimensional vectors (representations). 

 We denote $\vh_v \in \sR^d$ and $\vr \in \sR^d$ the representations for node $v$ and relation $r$, respectively. We denote $\gN_e(v)$ and $\gN_r(v)$ the set of node $v$'s neighboring entities and relations (including self-links), respectively. For example, $v' \in \gN_e(v)$ and $r \in \gN_r(v)$ if a fact $(v',r,v)$ exists. We use the terms nodes and entities interchangeably.

\subsection{KGQA}\label{sec:kgqa-back} 

Given a KG $\gG := (\gV, \gR, \gF)$ and a natural language question $q$, the task of KGQA is to extract a set of entities $\{a\} \in \gV$ that correctly answer $q$. 
Following the standard setting in KGQA~\cite{sun2018open}, we assume that the entities referred in the question are given and linked to nodes of $\gG$ via entity linking algorithms~\cite{yih2015semantic}. We denote these entities as $\{e\}_q \in \gV$ (seed entities),  e.g., \textit{Q. Tarantino} in \Figref{fig:ex}. 

The problem complexity is further reduced by extracting  a question-specific subgraph $\gG_q:= (\gV_q, \gR_q, \gF_q) \subset \gG$ which is likely to contain the answers (more details in Section~\ref{sec:exper-data}).
Each question $q$ and its answers $\{a\}_q \in \gV_q$, referred to as question-answer pair, induces a question-specific labeling of the nodes. Node $v \in \gG_q$ has label $y_v =1$ if $v \in \{a\}_q$ and $y_v = 0$ otherwise. The task can be thus reduced to performing binary node classification over $\gG_q$. 

The KGQA problem involves two modules: (i) retrieving a question-specific $\gG_q$ and (ii) reasoning over $\gG_q$ to perform answer classification. In this work, we introduce a new way to advance KGQA reasoning capabilities.

\subsection{GNNs}
GNNs~\cite{kipf2016semi,schlichtkrull2018modeling}  are well-established graph representation learners suited for tasks such as node classification. Following the message passing strategy~\cite{gilmer2017neural}, the core idea of GNNs is to update the representation of each node by aggregating itself and its neighbors’ representations. 

The GNN updates node representation $\vh_v^{(l)}$ at layer $l$ as
\begin{equation}
            \vh_v^{(l)} = \psi \Big( {\vh}_{v}^{(l-1)}, \phi \big( \{ \vm^{(l)}_{v'v}: v' \in \gN_e(v) \big) \Big),
            \label{eq:gnn0}
        \end{equation}
where $\vm_{v'v}^{(l)}$ is the message between two neighbors $v$ and $v'$, and $\phi(\cdot)$ is an aggregation function of all neighboring messages. Function $\psi(\cdot)$ combines representations of consecutive layers.
At each layer, GNNs capture 1-hop information (neighboring messages). An $L$-layer GNN model captures the neighborhood structure and semantics within $L$ hops.

\subsection{GNNs for KGQA} \label{sec:gnn-kgqa-back}
To better reason over multiple facts (graphs), successful KGQA methods utilize GNNs~\cite{sun2018open, he2021improving}. The idea is to condition the message passing of \Eqref{eq:gnn0} to the given question $q$. For example, if a question refers to movies, then 1-hop movie entities are more important. It is common practice~\cite{qiu2020stepwise, he2021improving, shi2021transfernet, lan2021survey} to decompose $q$ into $L$ representations $\{\vi^{(l)}\}^L_{l=1}$ (instructions), where each one may refer to a specific question's context, e.g., movies or actors. 

The instructions are used to guide different reasoning steps over $\gG_q$ by writing the GNN updates as
    %\begin{enumerate}
        %\item Initialize node representation $\vh_v^{(0)}$, for all $v \in \gG_q$.
        %\item For $l=1, \dots, L$, update node representations as
        \begin{equation}
            \vh_v^{(l)} = \psi \Big( {\vh}_{v}^{(l-1)}, \phi \big( \{ \vm^{(l)}_{v'v}: v' \in \gN_e(v) | \vi^{(l)}  \}\big) \Big),
            \label{eq:gnn}
        \end{equation}
        where each GNN layer $l$ is now conditioned to a different instruction $\vi^{(l)}$. Message $\vm^{(l)}_{v'v}$ usually depends on the representations of the corresponding fact $(v', r, v)$. 
        
The goal of GNNs is to selectively aggregate information from the question-relevant facts. Via \Eqref{eq:gnn}, GNNs learn to match each $\vi^{(l)}$ with 1-hop neighboring facts. Using the instructions $\{\vi^{(l)}\}_{l=1}^L$ recursively, GNNs learn the sequence of facts (KG traversal) that leads to the final answers. 

% For example, the question ``\textit{Who are the actors of the movies directed by Quentin Tarrantino?}'' corresponds to a ground-truth reasoning paths, such as \textit{(Quentin Tarrantino, directed, Pulp Fiction)}$\xrightarrow{}$\textit{(Pulp Fiction, actors, Uma Thurman)}.

% Motivated by this, state-of-the-art methods ~\cite{sun2019pullnet, he2021improving, atzeni2021sqaler} have additional mechanisms to identify the question-relevant sequences of  entities and relations that need to visited during KGQA. This is proven to be particularly helpful when the question corresponds to a composition of different KG relations (multi-hop questions)~\cite{atzeni2021sqaler}.

\begin{figure*}[t]
\centering
\resizebox{0.9\linewidth}{!}{\input{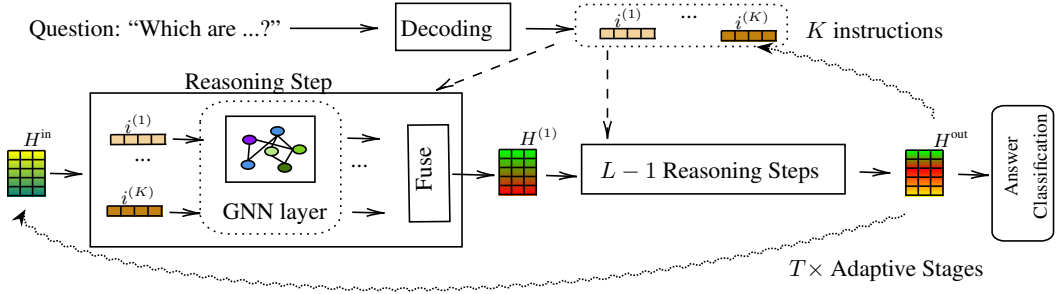}}
\caption{\method's adaptive reasoning. The question is decoded to $K$ instructions. At $L$ reasoning steps, we perform a BFS execution of the instructions. The procedure is repeated for $T$ adaptive stages that enhance the initial instruction decoding. }
\label{fig:frame}
\end{figure*}

\section{\method Approach}

\method (Reason \& Revise) enhances instruction decoding and execution for  effective KGQA reasoning. Our contributions across these two dimensions are described below.

\subsection{\method's BFS Instruction Execution} \label{sec:execute}

GNN updates in  \Eqref{eq:gnn} execute the instructions in a pre-defined order, which assumes that the generated instruction sequence matches exactly the information that is present in the KG. \method does not impose this assumption. Instead, it improves instruction execution by selecting the order by which to process the instructions on the fly, e.g., based on the KG semantics.

To achieve this, we emulate the breadth-first search strategy (BFS) by modifying the GNN updates in  \Eqref{eq:gnn}. The idea is to reason with all instructions at each step (breadh-first), before we decide which execution results to accept. We decompose the question into $K$ instructions $\{\vi^{(k)}\}_{k=1}^K$, but the number of instructions $K$ is now decoupled form the number of GNN layers $L$. We derive instruction-specific representations $\tilde{\vh}_v^{(k,l)}$ at each GNN layer as
\begin{equation}
    \tilde{\vh}_v^{(k,l)} = \phi \big( \{ \vm^{(l)}_{v'v}: v' \in \gN_e(v) | \vi^{(k)}  \}\big).
    \label{eq:gnn-ins}
\end{equation}
To allow the model select which instruction-specific representations are useful, we fuse them with a learnable function $\psi(\cdot)$ as
\begin{equation}
    \vh_v^{(l)} = \psi \big( \vh_v^{(l-1)}, \{ \tilde{\vh}_v^{(k,l)} \}_{k=1}^K \big).
    \label{eq:gnn-fuse}
\end{equation}
As a result, the reasoning module has the capability to decide on the final execution plan, i.e., which instructions are accepted at which GNN layers.

Specifically, we compute message $\vm_{v'v}^{(l)}$ between nodes $v'$ and $v$ as the representation of their corresponding relation $\vr_{v'v} \in \sR^{d}$ followed by a learnable projection matrix $\mW_R^{(l)} \in \sR^{d \times d}$. We condition $\vm_{v'v}^{(l)}$ to the underlying instruction $\vi^{(k)} $ by an element-wise multiplication followed by ReLU$(\cdot)$ nonlinerarity. In summary, we obtain
\begin{equation}
    \vc_{v'v}^{(k,l)} = \text{ReLU}( \vi^{(k)} \odot \mW_R^{(l)} \vr_{v'v}),
\end{equation}
where $\vc_{v'v}^{(k,l)}$ denotes the question-relevant message from node $v'$ at layer $l$ and for instruction $k$.

To measure the importance of $\vc_{v'v}^{(k,l)}$from node $v'$, we multiply it with  $p_{v'}^{(l-1)}$ of the previous layer, where $\vp^{(l)} \in [0,1]^{|\gV_q|}$ is a probability vector (we shortly describe how it is computed). We aggregate neighboring facts for node $v$ with a sum-operation and  \Eqref{eq:gnn-ins} becomes
\begin{equation}
    \tilde{\vh}_v^{(k,l)} = \sum_{v' \in \gN(v)} p_{v'}^{(l-1)} \vc_{v'v}^{(k,l)}.
\end{equation}

To combine node representations from different instructions, we use the column-wise concatenation operation $||$ followed by a learnable projection matrix  $\mW_h^{(l)} \in \sR^{d \times (K+1)d}$ (we observe similar performance with other fusion mechanisms, such as self-attention).  \Eqref{eq:gnn-fuse} becomes
\begin{align}
    \tilde{\vh}_v^{(l)} &= \big|\big|_{k=1}^K \tilde{\vh}_v^{(k,l)} \\
    \vh_v^{(l)} & = \text{ReLU} \big( \mW_h^{(l)} (  \vh_v^{(l-1)} || \tilde{\vh}_v^{(l)}) \big).
    \label{eq:psi}
\end{align}
Finally, we collect $\vh_v^{(l)}$ for all nodes to $\mH^{(l)}$ and compute the probability vector $\vp^{(l)}$ as
\begin{equation}
    \vp^{(l)} = \softmax(  {\mH^{(l)}} \vw ).
    \label{eq:pl}
\end{equation}
Initially, we set $p^{(0)}_v = 0$ if $v \notin \{e\}_q$  and $p^{(0)}_v = 1$ otherwise,  so that we start reasoning from the seed entities.

\subsection{\method's Adaptive Instruction Decoding} \label{sec:decode}

If the instructions are computed based solely on the question's context, they are not conditioned with respect to the underlying KG. This is important when we need to reason over multiple facts (complex questions), over KGs with rich semantics (see \Figref{fig:ex}) or over missing information~\cite{sun2018open}. 
% To improve instruction decoding, we repeat the reasoning process in an adaptive manner, where we iteratively condition the instructions with respect to the underlying KG semantics. 

\Figref{fig:frame} shows how our adaptive reasoning works. After reasoning with $L$ steps via \Eqref{eq:gnn-ins} and \Eqref{eq:gnn-fuse}, we keep the reasoning output (node representations $\mH^\text{out}$) that contains KG-aware information. We use it to update the initial instruction decomposition $\{\vi^{(k)}\}_{k=1}^K$ as well as the initial node representations $\mH^\text{in}$. This procedure is repeated for $T$ adaptation stages. Our goal is twofold; (i) to ground the instruction decomposition to underlying KG semantics, and (ii) to guide the reasoning process in an iterative manner. For example, some instructions may correspond to a part of the question that needs to be answered first, before reasoning with the rest instructions. 

To update each $\vi^{(k)}$ at every stage $t \in \{1, \dots, T\}$, we use the seed entities' final representations as the KG-aware information, which are computed by
\begin{equation}
\vh_e = \sum_{v \in \{e\}_q} \vh^{(L)}_v,
\end{equation}
and compute the adapted instructions $\vi^{(k)}$ as
\begin{align}
    \vi^{(k)} =& (\mathbf{1}-\vg^{(k)} ) \odot  \vi^{(k)} + \nonumber \\
    &\vg^{(k)} \odot  \mW_q (\vi^{(k)} || \vh_e ||  \vi^{(k)}-\vh_e ||  \vi^{(k)} \odot \vh_e). \label{eq:upd}
\end{align}
Here, $\mW_q \in \sR^{d \times 4d}$ are learnable parameters and  $\vg^{(k)} \in [0,1]^d$ is the output gate vector computed by a standard GRU~\cite{cho2014learning}. At every reasoning stage $t$, we  set $\mH^\text{in} = \mH^{(L)}$ to encode information of the previous stage. However, we reset the probability vector $\vp^{(0)}$ to the seed entities.

The algorithmic procedure of our \method method is summarized in Algorithm~\ref{alg:method}. It takes as input a question $q$ with seed entities  $\{e\}_q$, and the corresponding question-specific KG subgraph $\gG_q$ and classifies nodes as answers or non-answers. The number of adaptation stages $T$, instructions $K$, and reasoning steps $L$ are hyper-parameters.

\begin{algorithm}[t]
\caption{The high-level algorithmic procedure of \method (inference).} \label{alg:method}
\begin{algorithmic}[1]
   \STATE {\bfseries Input:} Question $q$, KG subgraph $\gG_q$, seed entities $\{e\}_q$, hyper-parameters: $T,K,L$.
   \STATE {\bfseries Initialize:} $K$ instructions $\{\vi^{(k)}\}_{k=1}^K$,  node representations ${\mH}^{\text{in}}$.
   \FOR{$t=1$ {\bfseries to} $T$} 
   %\COMMENT
   %\IF{\textcolor{blue}{BFS}}
    \FOR{$l=1$ {\bfseries to} $L$}
        \FOR{$k=1$ {\bfseries to} $K$}
            \STATE $\tilde{\vh}_v^{(k,l)} = \phi \big( \{ \vm^{(l)}_{v'v}: v' \in \gN(v) | \vi^{(k)}  \}\big)$.
             
           \ENDFOR
           \STATE $\vh_v^{(l)} = \psi \big( {\vh}^{(l-1)}_{v}, \{\tilde{\vh}^{(k,l)}_v\}_{k=1}^K \big)$.
       %\STATE Update  $\vh_v^{(k+1, 0)}$.
       \ENDFOR
       \STATE Set $\mH^\text{out}$ and $\mH^\text{in}$ to $\mH^{(L)}$.
       \STATE Update $\{\vi^{(k)}\}_{k=1}^K$ using ${\mH}^{\text{out}}$.
    %   \STATE Generate feedback $\vf_q = \text{FEED}(\vh_e^{(L)}, \tilde{\vs}^{(L)})$. 
    %   \STATE Update $\vi^{(k)} = \text{UPD}(\vi^{(k)} | \vf_q)$, $\forall k \in K$.
    \ENDFOR
    \STATE {\bfseries Result:} Classify node $v$ as answer or non-answer based on $\vh^\text{out}_v$.
\end{algorithmic}
\end{algorithm}

\subsubsection{Optimization and Initialization}
After reasoning for $T$ stages, we obtain the final probability vector $\vp^\text{out}$ by applying \Eqref{eq:pl} to $\mH^\text{out} = \mH^{(L)}$. Here, we omit the superscript $t \in \{1, \dots, T\}$ for readability, e.g., $\mH^{(T,L)}$.
We optimize the model's parameters with a classification based loss function (we use the KL-divergence~\cite{kullback1951information}), so that $p_v^\text{out}$ is close to 1 if $v \in \{a\}_q$ and zero otherwise. During inference, since we do not have the answer nodes $\{a\}_q$, we rank the nodes as possible answers based on their final probabilities.

For better generalization to unobserved entities, we initialize node representations $\vh^\text{in}_v $ as a function of their direct relations $r  \in \gN_r(v)$, as
\begin{equation}
    \vh^\text{in}_v = \text{ReLU} \big( \sum_{r \in \gN_r(v)}  \mW_0 \vr \big),
\end{equation}
where $\vr \in \sR^d$ is the representation of relation $r$ and $\mW_0 \in \sR^{d \times d}$ are learnable parameters. We derive the representation $\vr$ with the same pre-trained language model used for the question based on the relation's surface form, if applicable. Otherwise, we randomly initialize and  update $\vr$ during training.

To capture multiple question's contexts, each instruction is initialized by dynamically attending to different question's tokens~\cite{qiu2020stepwise,he2021improving}. First, we derive a representation $\vb_j$ for token $j$ and a question representation $\vq$ with pre-trained language models, such as SentenceBERT~\cite{reimers2019sentence} (see also Appendix). Each instruction $\vi^{(k)}$ is computed by
\begin{equation} 
\vi^{(k)} =  \sum_j u^{(k)}_j \vb_j,  
\end{equation}
where $u^{(k)}_j \in [0,1]$ is an attention weight for token~$j$. To ensure different instructions can attend to different tokens, the dynamic attention is computed by
\begin{align} 
 u^{(k)}_j &=  \softmax_j (\mW_u (\vq^{(k)} \odot \vb_j) ), \nonumber\\
 \vq^{(k)} &= \mW^{(k)} (\vi^{(k-1)}|| \vq || \vq \odot \vi^{(k-1)} || \vq - \vi^{(k-1)}),
\end{align}
where $\mW_u \in \sR^{d \times d}$ and $\mW^{(k)} \in \sR^{d \times 4d}$ are learnable parameters.

\subsection{Complexity}
\method's  computational complexity for each question is $\gO(T|\gV_q|L\Delta)$, assuming the KG is sparse with maximum node degree equal to $\Delta$. We also assume that the inner loop in Algorithm~\ref{alg:method} (line 6) is parallelized.
On the other hand, the computations of traditional GNN-based KGQA methods (\Secref{sec:gnn-kgqa-back}) can be achieved in $\gO(|\gV_q|L\Delta)$ time. However, we find that having $T=2$ is sufficient for \method in practice, so \method's complexity does not necessarily increase linearly.

\section{Experimental Setup}\label{sec:exper-data}
We experiment with three widely used KGQA benchmarks: WebQuestionsSP (Webqsp)~\cite{yih2015semantic}, 
Complex WebQuestions 1.1 (CWQ)~\cite{talmor2018web}, and 
MetaQA-3~\cite{zhang2018variational}. 
We provide the final dataset statistics (see \Secref{sec:implem}) in Appendix.
%\subsection{Datasets}

\subsection{Dataset Details}
\textbf{Webqsp} contains 4,737 natural language questions that are answerable using a subset Freebase KG. This  KG contains 164.6 million facts and 24.9 million entities. The questions require up to 2-hop reasoning within this KG. Specifically, the model needs to aggregate over two KG facts for 30\% of the questions, to reason over constraints for 7\% of the questions, and to use a single KG fact for the rest of the questions.
\noindent

\textbf{CWQ} is generated from Webqsp by extending the question entities or
adding constraints to answers, in order to construct  more complex multi-hop questions (34,689 in total). There are
four types of question: composition (45\%), conjunction (45\%), comparative (5\%), and superlative
(5\%). The questions require up to 4-hops of reasoning over the KG, which is the same KG with Webqsp.

\textbf{MetaQA-3} consists of more than 100k 3-hop questions in the domain of movies. The questions were constructed using the KG provided by the WikiMovies~\cite{miller2016key} dataset, with about 43k entities and 135k triples.

% \textbf{Subgraph Extraction}. For extracting question-specific subgraphs $\gG_q$, we follow the same procedure as previous works~\cite{sun2018open,sun2019pullnet,he2021improving}. First, we run the PageRank-Nibble~\cite{andersen2006local} (PRN) method (random walks with restart) from the seed entities to score other entities to be included in the subgraph. We set $\epsilon=1e^{-6}$ for PRN and select the top $m$ entities, where $m=2,000$ for Webqsp (full KG) and CWQ, and $m=500$ for MetaQA-3 and Webqsp (incomplete KG). The overall recall of answers among the subgraphs is 94.9\% for Webqsp, 79.3\% for CWQ and 99.0\% for MetaQA-3. 

\subsection{Implementation and Evaluation Details}\label{sec:implem}

Recall that our \method takes as input the question's seed entities $\{e\}_q$ and a question-specific subgraph $\gG_q$.
We use the seed entities provided by~\cite{yih2015semantic} for Webqsp, by~\cite{talmor2018web} for CWQ, and by~\cite{miller2016key} for MetaQA-3. We obtain subgraphs by ~\cite{he2021improving}. It runs the PageRank-Nibble~\cite{andersen2006local} (PRN) method from the seed entities to select the top-$m$ entiites to be included in the subgraph, as in~\cite{sun2018open}. We have $m=2,000$ for Webqsp (full KG) and CWQ, and $m=500$ for MetaQA-3 and Webqsp (incomplete KG). 

We tune the hyper-parameters $T$ (number of iterations), $K$ (number of instructions), and $L$ (number of GNN layers) amongst $T \in \{2,3\}$, $K \in \{2,3\}$, and $L \in \{2,3,4\}$. We perform model selection based on the best validation scores (more implementation details in the Appendix). 
For evaluation, we adopt two widely used metrics, Hits@1, which is the accuracy of the top-predicted answer, and the F1 score. To compute the F1 score we set a threshold equal to 0.95. For the competing approaches, we reuse the evaluation results reported in the corresponding papers, unless otherwise stated.

\subsection{Competing Approaches}
% We compare with methods that focus on improving KGQA reasoning capabilities. reasoning-based  and retrieval-based   approaches that solve weakly-supervised KGQA. Unless otherwise stated, we reuse the evaluation results reported in the corresponding papers.

We compare with methods that focus on improving KGQA reasoning capabilities. KV-Mem~\cite{miller2016key} is a key-value memory network~\cite{sukhbaatar2015end} for KGQA. EmbedKGQA~\cite{saxena2020improving} utilizes KG pre-trained embeddings~\cite{trouillon2016complex} to improve multi-hop reasoning. 
GraftNet~\cite{sun2018open}, HGCN~\cite{han2020open} and SGReader~\cite{xiong2019improving} are GNN-based approaches, where GraftNet and HGCN use a convolution-based GNNs~\cite{kipf2016semi}, while SGReader uses attention-based GNNs~\cite{velivckovic2017graph}. 

NSM~\cite{he2021improving} is the adaptation of Neural State Machines~\cite{hudson2019learning} to KGQA and performs a GNN-based reasoning. NSM-distill~\cite{he2021improving} improves NSM for multi-hop reasoning by learning which intermediate nodes to visit via distillation~\cite{hinton2015distilling}. TransferNet~\cite{shi2021transfernet} improves multi-hop reasoning over the relation set. EmQL~\cite{sun2020faithful} and Rigel~\cite{sen2021expanding} improve the ReifiedKB~\cite{cohen2020scalable} scalable baseline for deductive reasoning and reasoning with complex questions, respectively.

In addition, we compare with methods that focus on improving the question-specific input subgraph $\gG_q$. PullNet~\cite{sun2019pullnet} is built on top of GraftNet, but learns which nodes to retrieve via selecting shortest paths to the answers. SQALER~\cite{atzeni2021sqaler} learns which relations (facts) to retrieve during KGQA by reconstructing KG traversals to answers.

\section{Experimental Results}\label{sec:exper-res}

\begin{table}[tb]
	\centering
	\caption{Performance comparison of different methods (Hits@1 or F1 scores in \%). Bold fonts denote the best methods. }
	\label{tab:res}%
	%\begin{small}
	\resizebox{\columnwidth}{!}{
	\begin{threeparttable}
		\begin{tabular}{@{}lcc@{}}
			%\cline{3-7}
			\toprule
			% &\multicolumn{2}{c}{Webqsp}&\multicolumn{2}{c}{MetaQA-1}&\multicolumn{2}{c}{MetaQA-2}&\multicolumn{2}{c}{MetaQA-3}&\multicolumn{2}{c}{CWQ}\\
			 & \textbf{Webqsp} & \textbf{CWQ} \\
			 Method & H@1 / F1 &  H@1   \\
			% Models&Hits&F1&Hits&F1&Hits&F1&Hits&F1&Hits&F1\\
			\midrule
			KV-Mem~\cite{miller2016key} &	46.7 / 38.6& 21.1 \\
			SGReader~\cite{xiong2019improving} & 67.2 / 57.3  & --  \\
			EmbedKGQA~\cite{saxena2020improving} &	66.6 / \; \ -- \; & * \\
			GraftNet~\cite{sun2018open} & 66.7 / 62.4  & 32.8   \\
			PullNet~\cite{sun2019pullnet} & 68.1 / \; \ -- \; & 45.9 \\
			TransferNet~\cite{shi2021transfernet} & 71.4 /  \;  \ -- \; & 48.6  \\
			Rigel~\cite{sen2021expanding} & 73.3 /  \; \ -- \; & 48.7  \\
			NSM~\cite{he2021improving}	&	68.7 / 62.8 & 47.6  \\
			NSM-distill~\cite{he2021improving} & 74.3 / 67.4 & 48.8  \\
			EmQL~\cite{sun2020faithful} & 75.5 /  \; \  -- \;   & *  \\
			SQALER~\cite{atzeni2021sqaler} & 70.6 / \; \ -- \;   & *  \\
			SQALER+GNN~\cite{atzeni2021sqaler} & 76.1 / \; \ -- \;   & *  \\
			\midrule
			\textbf{\method} & \textbf{76.4} / \textbf{70.9} &  \textbf{52.9}  \\
			\bottomrule
		\end{tabular}%
		\begin{tablenotes}
		\item --: Result not reported.
		\item *: Method cannot inherently tackle this setting.
        \end{tablenotes}
		\end{threeparttable}
}

	%\end{small}
\end{table}%

%%%
% model dfs: webqsp: 72.1/67.3

% & Webqsp& CWQ &  MetaQA-2 & MetaQA-3 \\
% 			Models & H@1 / F1 &  H@1 / F1 & H@1 & H@1 \\
% 			% Models&Hits&F1&Hits&F1&Hits&F1&Hits&F1&Hits&F1\\
% 			\midrule
% 			KV-Mem&	46.7 / 38.6& 21.1 / \ \ \ -- &  82.7& 48.9\\
% 			KAReader & 67.2 / 57.3  & -- &  -- & -- \\
% 			GraftNet & 66.7 / 62.4  & 32.8 / \ \ \ -- &  94.8& 77.7 \\
% 			PullNet & 68.1 / \; -- \; & 45.9 / \ \ \ -- &  \textbf{99.9}& 91.4 \\
% 			SRN & -- &  -- &  95.1& 75.2\\
% 			EmbedKGQA &	66.6 / \; -- \; & -- &  98.8 & 94.8\\
% 			NSM	&	68.7 / 62.8 & 47.6 / 42.4 &  \textbf{99.9}& \textbf{98.9}\\
% 			NSM-distill & \textbf{74.3} / 67.4 & 48.8 / 44.0  & \textbf{99.9}& \textbf{98.9}\\
% 			\hline
% 			\textbf{BFS} & &  &  &  \\
% 			\; \methodg & 69.5 / 64.6 & 48.3 / 45.6 &  99.8 & 97.5  \\
% 			\; \method & 73.3 / 68.9 & \textbf{50.4} / \textbf{46.3} &  \textbf{99.9} & \textbf{98.9} \\
% 			\textbf{DFS} & &  &  &  \\
% 			\; \methodg & 72.6 / 68.4 &  &  \textbf{99.9} & \textbf{98.9}  \\
% 			\; \method & 74.0 / \textbf{69.7} & 49.3 / 46.6 &  \textbf{99.9} & \textbf{98.9} \\

\subsection{Main Results}

We present the KGQA performance for the compared methods in Table~\ref{tab:res}.  \method outperforms the best performing  method by 0.3\% and 4.1\% points at H@1 for Webqsp and CWQ, respectively.  

For Webqsp, although most questions involve one-hop or two-hop reasoning, few training examples are given. Methods such as NSM-distill and SQALER+GNN tackle this challenge with additional supervision signals (compare NSM with NSM-distill and SQALER with SQALER-GNN), while EmQL uses pre-defined question templates to facilitate instruction execution. In contrast, \method relies on its adaptive reasoning. If we compare \method with other reasoning-based approaches (NSM, GraftNet, and SGReader), \method performs better by 5.7-9.7\% (H@1 points) and 8.1-13.6\% (F1 points).

In CWQ, many questions include multiple seed entities and require both composition (sequential) and conjunction (parallel) reasoning, which makes the instruction execution challenging. \method's breadth-first strategy and adaptive updates are designed to benefit such challenging cases. Note that some methods cannot inherently tackle this setting: EmbedKGQA requires single-entity questions, EmQL requires pre-defined question templates that are hard to derive for complex questions, and SQALER assumes that only composition question types are involved. \method outperforms all other methods by more than 4\% points at H@1.

\subsection{Low-Data Regime Results}
Webqsp contains mostly simple questions which are easily answerable over a full KG.
Table~\ref{tab:incomplete-results} shows the performance for a more challenging task, when keeping 10\%, 30\%, and 50\% of the KG’s total facts.
We compare against GraftNet, SGReader, and HGCN, which are GNN-based approaches especially designed for reasoning over incomplete  KG subgraphs.  \method outperforms competing methods by  1.1-5.7\% points at H@1 and by 0.7-5.6\% points at F1. The best improvement is obtained for KG-50\%, since there is more KG information that \method can leverage during its adaptive reasoning. 

 MetaQA-3 has more than 100k train questions which involve only few KG relations. For a more challenging setting, we experiment with MetaQA-3 when we decrease the ratio of the KG completeness as well as the number of training question-answer pairs. We compare against NSM and NSM-distill that also rely on subgraph extraction with PRN (\Secref{sec:implem}). We provide additional results in the Appendix.  Table~\ref{tab:meta} shows that the more challenging the setting is, the better the improvement \method achieves over NSM and NSM-distill. When the KG is 50\% complete and we only use 1\% of the training questions, \method improves over NSM-distill by more than 10\% points at H@1.

\begin{table}
\centering
\caption{H@1 / F1 results in \% for Webqsp with incomplete KGs. ``\% KG completeness'' denotes the percent of the remaining KG facts.}
\label{tab:incomplete-results}
\resizebox{\columnwidth}{!}{%
%\begin{threeparttable}
\begin{tabular}{lccc}
\toprule
% & \multicolumn{3}{c}{Webqsp}  \\
\% of KG completeness & \multicolumn{1}{c}{10\%} & \multicolumn{1}{c}{30\%} & \multicolumn{1}{c}{50\%} \\
\midrule
% Pullnet* & -- & 34.6 / -- &  47.4 / -- \\
% EmbedKGQA* & -- & 31.4 / --  &  42.5 / --  \\
% LEGO* & -- &\textbf{38.0}/ -- & 48.4 / --\\
% \hline
GraftNet~\cite{sun2018open} & 15.5 / 6.5 & 34.9 / 20.4 & 47.7 / 34.3 \\
%NSM-distill~\cite{he2021improving} & 16.8 / 7.2 & 33.7 / 21.2 & 48.9 / 35.7 \\
SGReader~\cite{xiong2019improving} & 17.1 / 7.0 & 35.9 / 20.2  &   49.2 / 33.5  \\
HGCN~\cite{han2020open} & 18.3 / 7.9 & 35.2 / 21.0 & 49.3 / 34.3  \\

%NSM & & 33.7 / 21.2 & 48.9 / 35.7 \\

\midrule
%\method  & &  &  \\
%\textbf{\method} (Glove+LSTM) & 18.6 / 8.3 & 36.5 / 22.3 & 50.3 / 37.8 \\
\textbf{\method} & \textbf{19.4} / \textbf{8.6} & \textbf{37.9} / \textbf{23.6} & \textbf{53.4} / \textbf{39.9} \\
%\; Deep BFS & & 35.4 / \textbf{22.9} & 49.4 / 37.4 \\
\bottomrule
\end{tabular}%

}
\end{table}

\begin{table}
\centering
\caption{H@1 results in \% for MetaQA-3 under different settings. ``\% KG completeness'' denotes the percent of the remaining KG facts and ``\% Train QAs'' the percent of training question-answer pairs used.}
\label{tab:meta}
\resizebox{0.9\columnwidth}{!}{%
\begin{threeparttable}
\begin{tabular}{@{}lrrrr@{}}
\toprule
 \multicolumn{1}{l}{\% of KG completeness}& 100\% & 100\%& 50\%  & 50\% \\
 \multicolumn{1}{l}{\% of Train QAs}& 10\% & 1\% & 10\% & 1\%\\
\midrule
NSM & 98.8 & 89.6 & 71.8 & 49.7 \\
NSM-distill & \textbf{98.9} & 98.2 &  72.3 & 51.5 \\
\midrule
\textbf{\method} & \textbf{98.9} & \textbf{98.6} & \textbf{75.4} &  \textbf{62.7} \\
\bottomrule
\end{tabular}%
% \begin{tablenotes}
% \end{tablenotes}
\end{threeparttable}
}

\end{table}

\subsection{Ablation Studies}

Table~\ref{tab:abla} verifies that KGQA
improvements stem from the algorithmic design of our method. For complex questions (CWQ), deriving the correct execution order of the instructions  becomes challenging. By treating the instructions as a set, our BFS execution provides a performance gain of 4.7\% points. When there are missing facts (Webqsp-50\%), grounding the instruction decoding to the available information becomes crucial. Our adaptive decoding leverages  this KG-aware information and provides a performance gain of 2.2\% points.

Moreover, we experiment with hyper-parameter sensitivity.  We intentionally set $L=2$ for MetaQA-3, although it requires 3-hop reasoning, and gradually increase $T \in \{1, \dots, 5\}$ to evaluate whether \method can reach the answers. For $T=4$ and $K \in \{2,3\}$, \method achieves 84.5\%-88.9\% at H@1. When we set $T=5$, \method further improves and achieves 98.7\%-98.8\% at H@1. In Table~\ref{tab:cases}, we provide two case studies  that show how the number $T$ of adaptive stages impacts answer retrieval.

In addition, we have performed the following
ablation study at MetaQA-3 as motivated by \Figref{fig:ex}. The idea is to switch some KG relations with semantically similar ones during inferece to evaluate \method's adaptiveness. We switch the KG relations \{\textit{directed by, written by, starred
actors, release year}\} (out of total 9 relations) to \{\textit{has executive, plot by, has cast, air on}\}
respectively. During training, we switch them with 5\% probability, but during testing, we switch them with a 
50\% or 100\% probability, which enforces a distribution shift over the underlying relations. \method
(T=2) with adaptive stages outperforms \method (T=1) without adaptive stages in both cases. It performs 87.3\% and 86.9\% at F1, while \method (T=1) performs 84.5\%
and 81.1\% at F1, respectively. This experiment also suggests that it is \method's algorithmic design that
leads to its improvements.

\begin{table}[tb]
	\centering
	\caption{H@1 results in \% (with performance drop) under different \method's modifications for Webqsp (50\% complete) and CWQ. BFS Execution is described in \Secref{sec:execute} and Adaptive Decoding in \Secref{sec:decode}.}
	\label{tab:abla}%
	%\begin{small}
	\resizebox{0.95\columnwidth}{!}{
		\begin{tabular}{@{}lll@{}}
		    
			%\cline{3-7}
			\toprule
			% &\multicolumn{2}{c}{Webqsp}&\multicolumn{2}{c}{MetaQA-1}&\multicolumn{2}{c}{MetaQA-2}&\multicolumn{2}{c}{MetaQA-3}&\multicolumn{2}{c}{CWQ}\\
			Modification  & \textbf{Webqsp-50\%} & \textbf{CWQ} \\
			 %&   H@1 / F1 &  H@1 \\
			% Models&Hits&F1&Hits&F1&Hits&F1&Hits&F1&Hits&F1\\
			\midrule
% 			\method & \textbf{53.4} / \textbf{39.9} & \textbf{52.9} \\
% 			\; wo. adapting ($T=1$) &  51.2 / 36.6 &  49.7  \\
% 			\; wo. BFS (sequential instructions)  & 52.6 / 37.8  & 48.2  \\
            \textbf{\method} & \textbf{53.4} & \textbf{52.9} \\
 			\; without BFS Execution & 52.6 (-0.8)  & 48.2 (-4.7) \\
 			\; without Adaptive Decoding &  51.2 (-2.2) &  50.5 (-2.4)  \\
			\bottomrule
		\end{tabular}%
		}
	%\end{small}
\end{table}%

\begin{table}[tb]
\centering
\caption{Question-answer pairs and predicted answers (with probabilities) with respect to the number $T$ of adaptive stages.}
\label{tab:cases}
\resizebox{\linewidth}{!}{%
\begin{threeparttable}
\begin{tabular}{l}
%Question & Answers & Predicted Answers (probabilities) \\
\toprule
 \textbf{Q}: Who wrote movies that share directors  with the movie \\
 The Comebacks? \textbf{A}: R. Schneider, T. Brady \\ 
 $T=3:$  2007 (0.99) \\
 $T=4:$  \textcolor{teal}{R. Schneider} (0.99) \\
 $T=5:$  \textcolor{teal}{R. Schneider} (0.5), \textcolor{teal}{T. Brady} (0.5) \\
 \hline
 \textbf{Q}: When did the movies release  whose writers  \\
also wrote Birdy? \textbf{A}: 1989 \\
 $T=3:$  \textcolor{teal}{1989} (0.60), GD. Goldberg (0.11) \\ 
 $T=4:$ \textcolor{teal}{1989} (0.99)\\
 $T=5:$ \textcolor{teal}{1989} (1.0) \\

\bottomrule
\end{tabular}%
% \begin{tablenotes}
% \end{tablenotes}
\end{threeparttable}
}

\end{table}

% \begin{table}[tb]
% \centering
% \caption{Question-answer pairs and predicted answers (with probabilities).}
% \label{tab:cases}
% \resizebox{\linewidth}{!}{%
% \begin{threeparttable}
% \begin{tabular}{ll}
% %Question & Answers & Predicted Answers (probabilities) \\
% \toprule
%  Q: Who wrote movies that share directors  & Q: When did the movies release \\
%   with the movie The Comebacks? & whose writers also wrote Birdy? \\
%  A: R. Schneider, T. Brady &  A: 1989\\
% \hline
% $T=3:$  2007 (0.99) & \textcolor{teal}{1989} (0.60), GD. Goldberg (0.11) \\
% $T=4:$  \textcolor{teal}{R. Schneider} (0.99) & \textcolor{teal}{1989} (0.99)\\
% $T=5:$  \textcolor{teal}{R. Schneider} (0.5), \textcolor{teal}{T. Brady} (0.5) & \textcolor{teal}{1989} (1.0)\\

% % \multirow{3}{*}{\parbox{4cm}{Q: when did the movies release whose writers also wrote ``Birdy" A: 1989 }} & $t=3:$ \textcolor{teal}{1989} (0.60), Gary David Goldberg (0.11) \\
% % &  $t=4:$ \textcolor{teal}{1989} (0.99) \\
% % &  $t=5:$  \textcolor{teal}{1989} (1.0) \\
% % \hline

% \bottomrule
% \end{tabular}%
% % \begin{tablenotes}
% % \end{tablenotes}
% \end{threeparttable}
% }

% \end{table}

%ModificationBFS-REAREVDFS-REAREVNone50.4 / 48.274.7 / 69.7wo. adaptive stepswo. global pooling72.3 / 67.6wo. both71.5 / 67.5

%webqsp No glob: BFS     DFS TEST F1: 0.6922, H1: 0.7389
% DFS
% TEST F1: 0.6922, H1: 0.7389
% TEST F1: 0.7005, H1: 0.7267

% BFS
% TEST F1: 0.6934, H1: 0.7383
% TEST F1: 0.6989, H1: 0.7285

\section{Conclusion}
Our method (\method) introduces a new way to KGQA reasoning with respect to instruction execution and decoding. We improve instruction decoding via adaptive reasoning, which updates the instructions with KG-aware information. We improve instruction execution by emulating the breadth-first search algorithm, which provides robustness to the initial instruction ordering. Experimental results
on three KGQA benchmarks demonstrate
the \method’s effectiveness compared with
previous state-of-the-art, especially when the
KG is incomplete or when we tackle complex
questions.

\section{Limitations}
Our contributions are on the reasoning part, and our
method assumes that we have the linked entities and a question-specific subgraph as input. Although improving entity linking is out of our scope, our approach cannot recover from entity linking errors. However, just like \method, all the methods that we compare
against make the same assumption and rely on external entity linking tools. The linked
entities are obtained as explained in \Secref{sec:implem}.

Moreover, our method assumes that using a single GNN layer per reasoning step (see \Figref{fig:frame}) is sufficient. This may not be the case when, for example, we need to reason for multiple steps with the same instruction. Our method could benefit from designing a multistep reasoning module.

\bibliography{kgqa}
\bibliographystyle{acl_natbib}

\clearpage
\appendix

\section{Appendix}
\label{sec:appendix}

\subsection{Dataset Statistics}
 In Table~\ref{tab:data}, we provide the dataset statistics with the subgraph extraction algorithm described in \Secref{sec:implem}.  We also include the 2-hop MetaQA-2~\cite{zhang2018variational} dataset, which is easier than MetaQA-3.

\begin{table}[h]
	\centering
	\caption{Datasets statistics. ``avg.$|\gV_q|$'' denotes average number of entities in subgraph, and “coverage” denotes the ratio of at least one answer in subgraph.}
	\label{tab:data}%
	%\begin{small}
	\resizebox{\columnwidth}{!}{
		\begin{tabular}{@{}l|rrr|rr@{}}
			%\cline{3-7}
			\toprule
			% &\multicolumn{2}{c}{Webqsp}&\multicolumn{2}{c}{MetaQA-1}&\multicolumn{2}{c}{MetaQA-2}&\multicolumn{2}{c}{MetaQA-3}&\multicolumn{2}{c}{CWQ}\\
			Datasets & Train & Dev  & Test & avg. $|\gV_q|$ & coverage (\%) \\
			
			% Models&Hits&F1&Hits&F1&Hits&F1&Hits&F1&Hits&F1\\
			\midrule
			Webqsp & 2,848 & 250 & 1,639 & 1,429.8 & 94.9\\
			CWQ &  27,639 & 3,519 & 3,531 & 1,305.8 & 64.4 \\
			MetaQA-2 &   118,980 & 14,872 & 14,872 & 469.8 & 100.0 \\
			MetaQA-3 &  114,196 & 14,274 & 14,274 & 497.9 & 99.0 \\
			\bottomrule
		\end{tabular}%
		}
	%\end{small}
\end{table}%

\subsection{Implementation Details}
To encode questions, we use SenteceBERT~\cite{reimers2019sentence}; although we observe similar results if we use BERT~\cite{devlin2018bert} or RoBERTa~\cite{liu2019roberta}. For MetaQA, we use an LSTM encoder~\cite{hochreiter1997long} and initialize tokens with Glove~\cite{pennington2014glove}.

The number of hidden dimensions $d$ is tuned amongst $\{50,100\}$. 
We optimize the model with Adam optimizer~\cite{kingma2014adam}, where the learning rate is set tuned amongst $\{1e^{-4}, 5e^{-4}, 1e^{-4}\}$ and the batch size is tuned amongst $\{8, 16, 40\}$. We tune the number of epochs amongst $\{10, 30, 50, 100,200\}$. We apply dropout regularization~\cite{srivastava2014dropout} with probability tuned amongst $\{0.1, 0.2, 0.3\}$. We perform model selection based on the best validation scores. For Webqsp, \method achieves a validation score of 78.4\% at H@1, and for CWQ, a validation score of 57.4\% at H@1.

We implemented \method using PyTorch~\cite{paszke2017automatic}, reusing the source code of~\cite{sun2018open,he2021improving}. Experiments were performed on a Nvidia Geforce RTX-2070 and on a Nvidia Geforce RTX-3090 GPU over 32GB and 128GB RAM machines. Our code is publicly available at \url{https://github.com/cmavro/ReaRev_KGQA}.

\subsection{MetaQA Results}
We provide MetaQA full results in Table~\ref{tab:meta-app}. We devide methods in two categories: (i) subgraph-based that may not be able to access all possible answers, and (ii) full-graph/retrieval-based that can access all KG facts. Subgraph-based methods performs worse when answers are missing (see MetaQA-3 and Table~\ref{tab:data}). 
\begin{table}[h]
	\centering
	\caption{H@1 performance comparison for MetaQA.}
	\label{tab:meta-app}%
	%\begin{small}
	\resizebox{\columnwidth}{!}{
	\begin{threeparttable}
		\begin{tabular}{@{}lcc@{}}
			%\cline{3-7}
			\toprule
			% &\multicolumn{2}{c}{Webqsp}&\multicolumn{2}{c}{MetaQA-1}&\multicolumn{2}{c}{MetaQA-2}&\multicolumn{2}{c}{MetaQA-3}&\multicolumn{2}{c}{CWQ}\\
			 & \textbf{MetaQA-2} & \textbf{MetaQA-3} \\
			 %& H@1 &  H@1 (answer recall)  \\
			% Models&Hits&F1&Hits&F1&Hits&F1&Hits&F1&Hits&F1\\
			\midrule
			\midrule
			Subgraph-based with PRN algorithm (\Secref{sec:implem}) & & \\
			\midrule
			GraftNet &  94.8 & 77.7 \\
			NSM	& \textbf{99.9} & \textbf{98.9}  \\
			NSM-distill & \textbf{99.9} & \textbf{98.9}  \\
			\textbf{\method} & \textbf{99.9} & \textbf{98.9}  \\
			\midrule
			\midrule
			Full graph/ Retrieval-based & & \\
			\midrule
			KV-Mem &	82.7 & 48.9 \\
			EmbedKGQA &	98.8 & 94.8 \\
			PullNet & 99.9 & 91.4\\
			
			EmQL & 98.6 & 99.1 \\
			SQALER & 99.9 & 99.9 \\
			SQALER+GNN & 99.9 & 99.9 \\
			TransferNet & \textbf{100} & \textbf{100}  \\
			\bottomrule
		\end{tabular}%
% 		\begin{tablenotes}
%         \end{tablenotes}
		\end{threeparttable}
}

	%\end{small}
\end{table}%

%%%
% model dfs: webqsp: 72.1/67.3

% & Webqsp& CWQ &  MetaQA-2 & MetaQA-3 \\
% 			Models & H@1 / F1 &  H@1 / F1 & H@1 & H@1 \\
% 			% Models&Hits&F1&Hits&F1&Hits&F1&Hits&F1&Hits&F1\\
% 			\midrule
% 			KV-Mem&	46.7 / 38.6& 21.1 / \ \ \ -- &  82.7& 48.9\\
% 			KAReader & 67.2 / 57.3  & -- &  -- & -- \\
% 			GraftNet & 66.7 / 62.4  & 32.8 / \ \ \ -- &  94.8& 77.7 \\
% 			PullNet & 68.1 / \; -- \; & 45.9 / \ \ \ -- &  \textbf{99.9}& 91.4 \\
% 			SRN & -- &  -- &  95.1& 75.2\\
% 			EmbedKGQA &	66.6 / \; -- \; & -- &  98.8 & 94.8\\
% 			NSM	&	68.7 / 62.8 & 47.6 / 42.4 &  \textbf{99.9}& \textbf{98.9}\\
% 			NSM-distill & \textbf{74.3} / 67.4 & 48.8 / 44.0  & \textbf{99.9}& \textbf{98.9}\\
% 			\hline
% 			\textbf{BFS} & &  &  &  \\
% 			\; \methodg & 69.5 / 64.6 & 48.3 / 45.6 &  99.8 & 97.5  \\
% 			\; \method & 73.3 / 68.9 & \textbf{50.4} / \textbf{46.3} &  \textbf{99.9} & \textbf{98.9} \\
% 			\textbf{DFS} & &  &  &  \\
% 			\; \methodg & 72.6 / 68.4 &  &  \textbf{99.9} & \textbf{98.9}  \\
% 			\; \method & 74.0 / \textbf{69.7} & 49.3 / 46.6 &  \textbf{99.9} & \textbf{98.9} \\
\end{document}